\documentclass[12pt]{article}

\usepackage{sbc-template}

\usepackage{graphicx,url}

\usepackage[latin1]{inputenc}  
\usepackage{amsfonts}
\usepackage{amsmath}
\usepackage{comment}
\usepackage{multirow}
\usepackage{float}
\usepackage{hyperref}

\usepackage[obeyFinal]{todonotes}
\usepackage[normalem]{ulem}

\title{Semantic Parsing Natural Language into SPARQL: Improving Target Language Representation with Neural Attention}
\author{Fabiano Ferreira Luz\inst{1}, Marcelo Finger\inst{1}}
\address{Universidade de São Paulo - USP\\
Instituto de Matemática e Estatística - IME\\
Departamento de Ciência da Computação\\
Rua do Matão, 1010 - Cidade Universitária\\
São Paulo - SP - Brasil - CEP 05508-090
\email{\{fluz,mfinger\}@ime.usp.br}
}

\begin{document} 

\maketitle

\begin{abstract}
Semantic parsing is the process of mapping a natural language sentence into a formal representation of its meaning. In this work we use the neural network approach to transform natural language sentence into a query to an ontology database in the SPARQL language. This method does not rely on handcraft-rules, high-quality lexicons, manually-built templates or other handmade complex structures. Our approach is based on vector space model and neural networks. The proposed model is based in two learning steps. The first step generates a vector representation for the sentence in natural language and SPARQL query. The second step uses this vector representation as input to a neural network (LSTM with attention mechanism) to generate a model able to encode natural language and decode SPARQL.
\end{abstract}
     
\section{Introduction}

Semantic parsing can be defined as the process of mapping natural language sentences into a machine interpreted, formal representation of its meaning. Currently, there are many efforts aimed at transforming human language into a computational representation~\cite{zettlemoyer2012learning,alshawi2014deterministic,bowman2014recursive}. 

In this work, we are concerned with semantic parsing as the task of transforming  natural language into SPARQL queries. In recent years this task has received a lot of attention~\cite{wang2007panto,lehmann2011autosparql,ferre2012squall, ferre2017sparklis}. This is mainly due to the increase of RDF-based documents available on the web, with public repositories such as DBPedia~\cite{auer2007dbpedia}, which already provides support for SPARQL queries in its database.

Traditional approaches to translating natural language into SPARQL rely on high-quality lexicons, manually-built templates, and linguistic features which are either domain or representation-specific \cite{dong2016language}. This is a problem since adapting such models to a new domain can be a very laborious task. Because of this, we propose a model based on artificial neural networks that among other advantages, does not rely on handcraft rules, high-quality lexicons, manually-built templates or other handmade complex structures.

In our work, we used an LSTM encoder-decoder model capable of encoding natural language (English) and decoding query language (SPARQL). The first part of our work is dedicated to finding suitable vector representations for sentences both in natural language and in SPARQL. On the natural language side, we use the vector-based Glove model~\cite{pennington2014glove}. For the SPARQL part, we propose a composition of methods to generate the vector representation. In addition, we propose our own way of generating this representation. In the second part of this work, we implement and configure an LSTM encoder-decoder model tailored to the task of translating natural language queries to SPARQL. 

The contributions of this paper are the following.

\begin{itemize}
\item The creation of a version of the Geo880 dataset with SPARQL queries like a target language.
\item The creation of an OWL ontology of the Geo880 domain.
\item The development of a novel vector representation for the target language lexicon.
\item The adaptation of the encoder-decoder model with neural attention to transform natural language into SPARQL queries.
\end{itemize}

This work is organized as follows. In Section ~\ref{sec:background}, we provide a background of the elements needed to understand how our model works. In Section ~\ref{sec:approach} we detail our model and the lexical generation procedure. Section ~\ref{sec:experimental} is dedicated to discussing experiments and results. In Section ~\ref{sec:relatedwork} we talk more about related work and in the last section we comment on our contribution and future work.

\section{Background}
\label{sec:background}

In the following sections we describe the neural network architecture encoder-decoder and also discuss the concept of neural attention.

\subsection{Recurrent Neural Network}

A Recurrent Neural Network (RNN) is a type of artificial neural network where connections between units form a directed cycle. This cycle represents an internal state of the network which allows it to exhibit dynamic temporal behavior. RNNs can use their internal memory to process arbitrary sequences of inputs. An output of a hidden layer $h_t$ of an RNN can be defined as 

\begin{align}
h_t&=f(h_{t-1},x_t),
\label{eq:encoder}
\end{align}

\noindent where $h_{t-1}$, is the value of the hidden layer at time $t-1$, $x_t$ is the input feature vector and $f(.)$ is a nonlinear function.

It has been noted by~\cite{bengio1994learning} that RNNs suffer from the \textit{vanishing gradient problem}, which consists of the exponential decrease that the value of $h_{t'}$ has influence over the value of $h_t$, $t' < t$, leading to a very short the temporal memory in the network. One solution to this problem was a change in the neuron's nucleus called Long Short Term Memory (LSTM)~\cite{hochreiter1997long}. RNN-LSTM has been used successfully in language modeling problems because, so now it can handle long sequences quite well.

\subsection{Encoder-Decoder Model}
The Encoder-Decoder model, proposed by \cite{cho2014learning}, is a neural network architecture that learns the conditional distribution of a conditioning variable-length sequence $\mathbf{x}$ in another variable-length sequence $\mathbf{y}$.  It performs this task by learning how to \textit{encode} a variable-length sequence $x_1,...,x_{T}$ into a fixed-length vector representation $c$ and then to \textit{decode} a given fixed-length vector representation $c$ back into a variable-length sequence $y_1,...,y_{S}$. The function may be interpreted as the distribution $p(y_1,...,y_{S} | x_1,...,x_{T})$; the input sequence length $T$ and output one $S$ can be different.

The \emph{encoder} is an RNN that reads each word of an input sequence $\mathbf{x}$ sequentially. As it reads each symbol, the hidden state of the RNN is updated according to equation~\eqref{eq:encoder}. After reading the end of the sequence (marked with an end sequence symbol), the hidden state of the RNN is summarized. We call this summary $c$. In order to simplify we can define $c$ as the output $h$.

\begin{figure}[H]
  \centering
  \includegraphics[width=.60\textwidth]{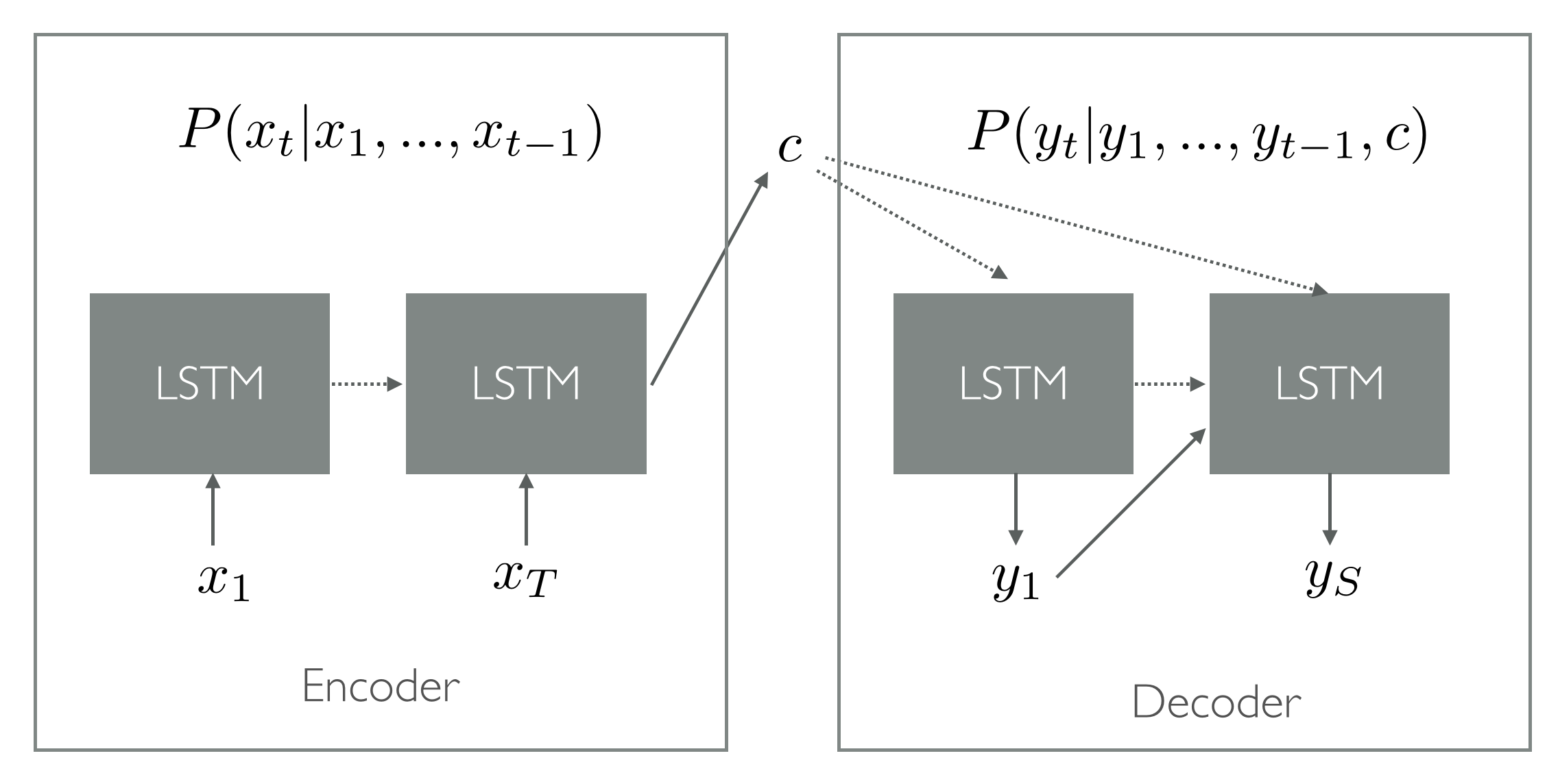} 
  \caption{Encoder-Decoder scheme. }
  \label{fig:enc-dec} 
\end{figure}

The \emph{decoder} is another RNN which is trained to generate the output sequence by predicting the next symbol $y_t$ given the hidden state $h_t$. However, unlike the RNN described previously, here both $y_t$ and $h_t$ are conditioned to $y_{t-1} $ and the summary $c$ of the input sequence. Thus, the hidden state of the decoder at time $t$ is computed by: $h_t=f(h_{t-1},y_{t-1},c)$, and likewise, we can define the conditional distribution of the next symbol by the following equation:
\begin{align}
\label{eq:prob-dec}
P(y_t|y_{t-1},...,y_1,c)=g(h_{t-1},y_{t},c).
\end{align}
The activation function $g$ produces valid probabilities by, for example, computing the \textit{softmax function}. Figure ~\ref{fig:enc-dec} presents an overview of the encoder and decoder scheme.

The combination of the two described components (Encoder and Decoder) make up the training of the proposed model to maximize the conditional log-likelihood and can be represented by the equation:

\begin{align}
\label{eq:train}
\max_{\theta} \frac{1}{N} \sum\limits_{n=1}^{N} \log p_{\theta}(\mathbf{y} | \mathbf{x}),
\end{align}

\noindent where $\theta$ is the set of the model parameters and each pair $(\mathbf{x},\mathbf{y})$
is, respectively, an input and output sequence. In our case, we use the vector representation of questions in natural language as input and the SPARQL query as an output. Since the output of the decoder, starting from the input, is differentiable, we can use a gradient-based algorithm to estimate the model parameters.

After training the encoder-decoder RNN, the model can be used in two distinct ways. In the first case, we can use the model to generate a target sequence, once the input sequence is provided computing the most probable output given the input. In the second one, the model can be used to evaluate a given pair of input and output sequences by calculating a score (e.g. the probability $p_\theta(\mathbf{y} | \mathbf{x})$). 

Although sequential models present good results for sequence transformation, they still present a distortion in 
the sentence mapping task because relationships among sentences are largely organized in terms of latent nested  syntactic/semantic structures rather than sequential surface order~\cite{lees1957syntactic,DyerKBS16}. One way to deal with these linguistic properties  is adding the mechanism of \textit{neural attention}, as used, for example, in soft alignment~\cite{Bahdanau2015}.

\subsection{Attention Mechanism in Neural Networks}


The model proposed by~\cite{Bahdanau2015} differs from the basic model of encoder-decoder by not attempting to encode a full entry into a fixed-size vector. Instead, it encodes entries into a sequence of vectors and selects a subset of them adaptively while decoding the translation. With this modification, the neural network no longer has the challenge of compressing information of an entire sequence into a fixed-size vector. The new architecture consists of a bidirectional RNN as an encoder and a decoder that emulates searching through a source sentence while decoding a translation. 

\paragraph{New Encoder:} The proposed encoder in \cite{Bahdanau2015} does not use a standard RNN described in equation \eqref{eq:encoder}, which reads an input sequence $\mathbf{x}$ starting from the first element $x_1$ to the last $x_{T_x}$. However, in the proposed scheme, the encoder does not compute only a single summary of the previous words. Instead, for each input word it computes an \textit{annotation} representing both a summary of the previous words and also one for the following ones. Then, an appropriate model to obtain such annotation is a bidirectional RNN~\cite{schuster1997bidirectional}, which has recently been used successfully in speech recognition~\cite{graves2013hybrid}.

A bidirectional RNN is composed of a forward RNN $\overrightarrow{f}$ that reads the input sequence as it is ordered from $x_1$ to $x_{T_x}$ and calculates a sequence of \textit{forward hidden states} $(\overrightarrow{h}_1,...,\overrightarrow{h}_{T_x})$ and the backward RNN $\overleftarrow{f}$ which reads the sequence in reverse order, from $x_{T_x}$ to $x_1$, resulting in a sequence of \textit{backward hidden states} $(\overleftarrow{h}_1,...,\overleftarrow{h}_{T_x})$.

Thus, an annotation is obtained for each word $x_j$ by concatenating the forward hidden state $\overrightarrow{h_j}$ and the backward one $\overleftarrow{h_j}$, 

\begin{equation}
h_j= \left[ \overrightarrow{h}_{j}^{T};\overleftarrow{h}_{j}^{T}\right]^{T} .
\end{equation}

The annotation $h_j$ encodes  both preceding and following words. Due to the tendency of RNNs to better represent recent inputs, the annotation $h_j$ is focused on the words around $x_j$. This sequence of annotations is used by the decoder and the alignment model later to compute the context vector.
\paragraph{New Decoder:} In the new model, each conditional probability is defined by: $p(y_i|y_1,...,y_{i-1},x) = g(y_{i},s_i,c_i)$, where $s_i$ is an RNN hidden state for time $i$, computed by:
\begin{align}
s_i&=f(s_{i-1},y_{i-1},c_i)
\label{eq:si}
\end{align}

\noindent Unlike the basic encoder-decoder approach in~\eqref{eq:prob-dec}, here the probability is conditioned on a distinct context vector $c_i$ for each target word $y_i$. The context vector $c_i$ depends on a sequence of \textit{annotations} $(h_1,...,h_{T_x})$ in which an encoder maps the input sentence. Each annotation $h_i$ consists of information about the whole input sequence with a strong focus on the parts surrounding the $i$-th word  of the input sequence.  

The context vector $c_i$ is then computed as a weighted sum of these annotations $h_i$

\begin{align}
c_i=\sum\limits_{j=1}^{T_x} \alpha_{ij} h_j
\end{align}
The weight $\alpha_{ij}$ of each annotation $h_j$ is computed by

\begin{equation}
\alpha_{ij}=\frac{\exp(e_{ij})}{\sum_{k=1}^{T_x}\exp(e_{ik})},
\end{equation}
\noindent where $e_{ij}=a(s_{i-1},h_j)$ is an alignment model that scores how well the input around position $j$ and the output at position $i$ match. The score is based on the RNN hidden state $s_{i-1}$ of~\eqref{eq:si} and the $j$-th annotation $h_j$ of the input sentence.

The alignment model $a$ is parametrized as a feedforward neural network which is concomitantly trained with all the other components of the proposed system. The alignment model directly computes a soft alignment, which allows the gradient of the cost function to be backpropagated. This gradient can be used to train the alignment model as well as the whole translation model at the same time.

The use of a weighted sum of all annotations can be interpreted as calculating an \textit{expected annotation}, where the expectation is over possible alignments. Let $\alpha_{ij}$ be a probability that the target word $y_i$ is aligned to, or translated from, a source word $x_j$. Then, the $i$-th context vector $c_i$ is the expected annotation over all the annotations with probabilities $\alpha_{ij}$. 

\begin{figure}[H]
  \centering
  \includegraphics[width=.40\textwidth]{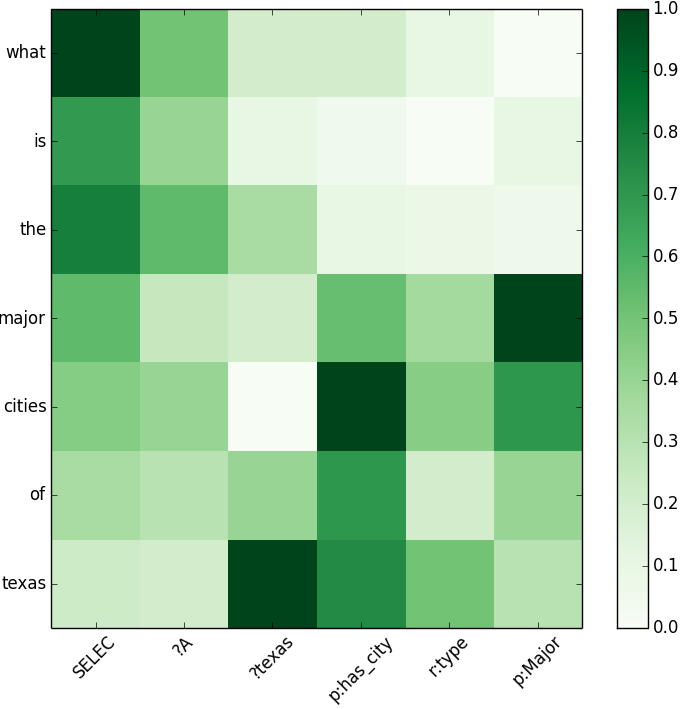} 
  \caption{Neural attention matrix example}
  \label{fig:attention} 
\end{figure}

For visual reasons, the alignment matrix of Figure ~\ref{fig:attention} was plotted with colors instead of numerical values. The matrix is an example of the alignment obtained during the training of our model. The darker the cell of the matrix the bigger the correlation between the terms row and column. 
\section{Our Approach}
\label{sec:approach}

The model described here is based on two approaches in the neural network literature. First, a neural probabilistic language model similar to that of~\cite{pennington2014glove} is used to learn a word vector representation, and a LSTM neural network similar to that of~\cite{dong2016language} is used to encode natural language sentences and decode SPARQL query. We provided the necessary background on these two components in Section ~\ref{sec:background}.

There are at least two important contributions in our work. The first is the fact that we use the model in question to translate from natural language to SPARQL. Another important contribution is the representation of the lexicon we are proposing. For this representation we use the neural attention mechanism to generate the table that will be used in matching. The present section is divided into three parts where, first, we give an overview of the developing model; then we describe the concept of \textit{matching} using neural attention; and in the end we detail how we generate the lexical representation for the target language.

\subsection{General overview of the approach}

Our approach consists of two learning phases, which can be observed in Figure ~\ref{fig:general}:

\begin{figure}[H]
  \centering
  \includegraphics[width=.60\textwidth]{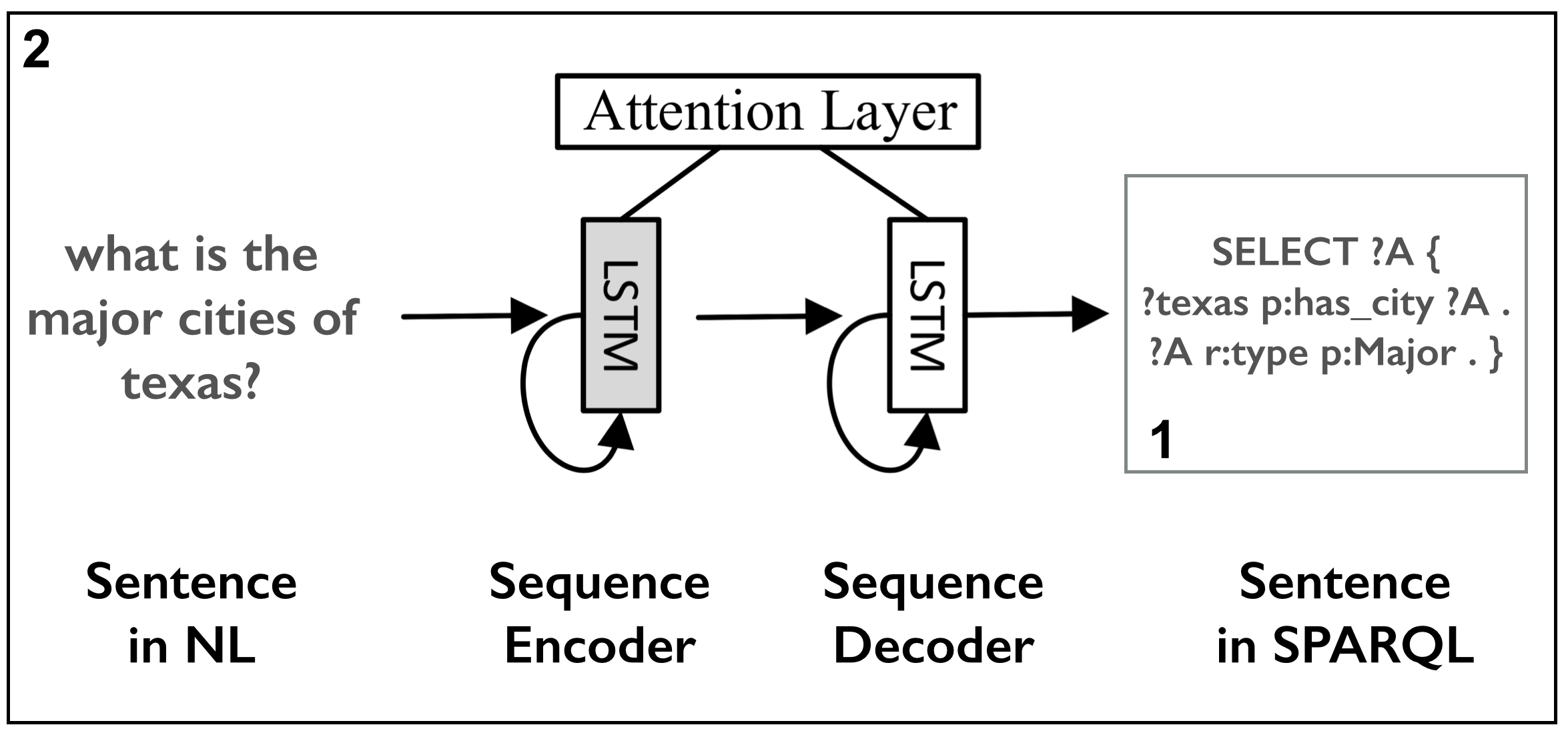} 
  \caption{Overview}
  \label{fig:general} 
\end{figure}

In the following text, we have a detailed description of each step:
\begin{enumerate}
\item The first step is to find a good vector representation for the target language lexicon. To represent the source natural language, we employ the method in \cite{pennington2014glove}.
\item In the second part, we are concerned with implementing and finding the settings so that our architecture can translate from natural language to SPARQL.
\end{enumerate}

For the training of our model, a dataset of paired sentences is necessary where, on the one hand we have questions in natural language and the other their SPARQL.  This dataset is described in Section~\ref{sec:experimental}.

\subsection{Vocabulary matching using neural attention}
\label{sec:maching}

In this work we propose a kind of dictionary where for each word of the target language we have another one in the source language that has the \textit{greatest correlation}. Let $T$ be target language vocabulary and $S$ be source language vocabulary. $\Lambda $ is the total alignment matrix of dimension $|S| \times |T|$  where $\Lambda_{ij}$ contains the value of correlation between words $s_i$ and $t_j$. We can define the \textit{greatest correlation} of a word $t_j$ from the target language as the:
\begin{equation*}
 \max\limits_{i=1..|S|} (\Lambda_{ij})
\end{equation*}
\noindent and the word $w_{i^*}$, $i^* = argmax(\Lambda_{ij})$, is the word from the source language that possesses the \textit{greatest correlation} with $t_j$.

We can exemplify the matching procedure with Figure~\ref{fig:attention} from the previous section. Note that the target word \textbf{p:has\_city} has the corresponding word \textbf{cities}. We can observe this because the darker cell connects the two words. That is, the word \textbf{p:has\_city} has \textit{greater correlation} with the word \textbf{cities}. Based on this table, we have built a translation dictionary from target to source. This dictionary could have several practical uses, for example, the lexical representation of the target. It will be better explained in the next section.

\subsection{Lexical representation}
\label{sec:lexicalrep}
To represent the lexicon of natural language, our source language, we use the model developed in \cite{pennington2014glove}. However, in the case of the SPARQL terms we did several experiments to analyze different representations. Next we describe the vector representation generation methods for the lexicon we use.
\vspace{-4mm}
\paragraph{RANDOM:} This representation was generated randomly using a normal distribution as a kernel with values from -1 to 1. This approach was chosen in order to kick off the representation like a baseline and even if it could not capture the relationship between terms, its performance was not so low when compared to more sophisticated approaches.
\vspace{-4mm}
\paragraph{TF-IDF/PCA:} We use the target language sentences and generate a term document matrix. In this matrix, we consider each query a document and each word in the target language a term. After generating the term-document matrix we apply TF-IDF \cite{aizawa2003information}. At the end we did a size reduction using Principal Component Analysis (PCA) \cite{jolliffe2002principal}. Our reduction was from 880 to 300 dimensions. 
\vspace{-4mm}
\paragraph{W2V10:} This approach is based on work by \cite{mikolov2010recurrent}, which proposed the generation of a vector representation of words based on recurrent neural networks, using only text as input. We use the target language sentences as input text to generate a vector representation. 
\vspace{-4mm}
\paragraph{OUR-APP:} Our approach is focused on using the same set of vectors to represent both source and target language vocabulary. What we do is a match between the terms of the source language and the target language seeking a match between the two vocabularies. The matching process is possible with the alignment table generated by the neural attention mechanism.

Then, first using a random version to represent the target's lexicon, we train and generate an alignment table, Figure~\ref{fig:attention}. Then, we use the \textbf{Matching} (see section~\ref{sec:maching}) to associate the lexicon of the source language with the target language. We use a heuristic to associate all elements of one vocabulary with another. SPARQL words are similar to English words, for example, ``SELECT'', ``FILTER'', ``ORDER'', etc were directly associated with their respective correspondents in English. The other terms such as, subject, predicate and object of the triples, are solved with the matching procedure. Our heuristics proved to be good according to the results. A full description of the heuristics used and how to deal with the generation of non-grammatical (incorrect) SPARQL expressions will be dealt with in future work.

Note that in this work we do not deal with the ``out of vocabulary''
situation (OOV), nor with lexical disambiguation problem.  In fact,
the OOV problem is avoided by the word2vec pre-processing, and every
word is assumed to be associated to a vector of features, which is
comparable to having  a controlled vocabulary.  Similarly, the lexical
ambiguities are assumed to be codified inside the attribute vectors in
the pre-processing; furthermore, approximately 1\% of the words in the
vocabulary are lexically ambiguous, allowing us to safely ignore such
effects.
 
\section{Experimental Evaluation}
\label{sec:experimental}

We compared our methodology with related work using the Geo880 dataset. We use only one dataset because we needed to create a SPARQL corpus for each domain and it is very laborious. The test with different datasets remained as future work.

In this section, first, we define our metrics and then talk about an adaptation of the dataset to the SPARQL query. We also talk about syntactic errors, neural network settings and finally comment on comparisons of our work with other different approaches.
Here we use two metrics to evaluate our approach:

\begin{scriptsize}
\begin{equation*}
\text{Accuracy} = \frac{ \text{\# of correctly translated queries}}{ \text{\# total of queries}}
\end{equation*}
\begin{equation*}
\text{Syntactical Errors} = \frac{ \text{\# of queries with syntactical errors}}{ \text{\# total of queries}}
\end{equation*}
\end{scriptsize}

\vspace{-10mm}
\subsection{Dataset}
Our experiments were conducted using one traditional dataset: Geo880, a set of 880 queries to a database of U.S. geography. The data were originally annotated with Prolog style semantics which we manually converted to equivalent statements in SPARQL queries. On the official web page of Geo880 dataset of the University of Austin in Texas\footnote{http://www.cs.utexas.edu/users/ml/nldata/geoquery.html} we found some files. Using two files, the \textbf{geobase} with assertions of the dataset and the \textbf{geoquery880} file, containing questions directed at this domain. First we created an OWL ontology based on the geodata file and then, for each query in natural language on geoquery880 file, we wrote a corresponding SPARQL query. Both the ontology and the set of questions can be found in our repository \url{https://github.com/mllovers/geo880-sparql} .

\subsection{Syntactical Errors}

We call a \emph{syntactical error} when a generated SPARQL statement cannot be processed by Protégé~\footnote{\url{http://protege.stanford.edu/}} due to syntactical formation\footnote{Another option would be to use a grammar that describes SPARQL, then use it as parameter of the parser algorithm. However a rejection by the SPARQL interpreter is a cheaper option found.}. In general, there are several syntactical errors that can be generated, such as not closing any brackets or even trying to apply a function to a variable of the wrong type.

\subsection{Settings}

Finding better parameters and hyper-parameters for neural networks is always a very costly task. The first rounds with the neural network, the pre-tests, served to find the best parameters for the neural network. In all, the pre-tests lasted more than two months.

\begin{itemize}
\item Learning rate: During the pre-test, we use two different learning rates. The first, we maintained the learning rate at 0.5. The second was started learning rate at 0.9 and decreasing 0.9 of it in each iteration (epoch). The best results obtained are used in the second case;
\item Epoch: In the tests we chose to use 100 epochs, as this was the best result found in our pre-tests for hyperparameters;
\item Hidden dimension: With regard to the number of hidden layers, we used a series of pre-tests with 100, 200 and 400 hidden layers. We continued testing with the 4 different dimensions;
\item Input dimension: We used three different input dimensions, 100, 200 and 300 but the best results were obtained with vectors of size 300.
\end{itemize}
\vspace{-3mm}
\subsection{Results}
\label{sec:results}
In Table~\ref{tab:lstm} we show the experiments performed without the mechanism of neural attention. All experiments were performed with the Geo880 dataset. We used 10-fold crossvalidation.

\vspace{-7mm}
\begin{tabular}{cc}
    \begin{minipage}{.4\linewidth}
\begin{table}[H]
{\tiny
\begin{center}
\begin{tabular}{ |l|l|r|r| }
\hline
\multicolumn{4}{ |c| }{LSTM Encoder-decoder - Geo880 Dataset} \\
\hline
Hidden Dim. & Method & Synt. error & Accuracy\\ \hline
\multirow{4}{*}{100} 
 & RANDOM 		& 14.77 & 30.68\\
 & TF-IDF/PCA & 13.64 & 31.89\\
 & W2V10 		& 14.77 & 28.40\\
 & OUR-APP 		& 11.36 & 38.63 \\ \hline
\multirow{3}{*}{200} 
 & RANDOM 		& 11.36 & 36.36\\
 & TF-IDF/PCA & 10.23 & 38.67\\
 & W2V10 		& 11.36 & 32.95\\
 & OUR-APP 		& 07.95 & 54.55 \\ \hline
\multirow{3}{*}{400} 
 & RANDOM 		& 09.09 & 51.14\\
 & TF-IDF/PCA & 07.95 & 54.55\\
 & W2V10 		& 10.23 & 42.04\\
 & \textbf{OUR-APP} & \textbf{06.81} & \textbf{64.77} \\ 
  \hline
\end{tabular}
\end{center}
\caption{Without neural attention}
\label{tab:lstm}
}
\end{table}

   \end{minipage} &
    \begin{minipage}{.4\linewidth}
\begin{table}[H]
{\tiny    
\begin{center}
\begin{tabular}{ |l|l|r|r| }
\hline
\multicolumn{4}{ |c| }{LSTM Encoder-Decoder - Geo880 Dataset} \\
\hline
Hidden Dim. & Method & Synt. Error & Accuracy\\ \hline
\multirow{4}{*}{100} 
 & RANDOM 		& 12.50 & 37.50\\
 & TF-IDF/PCA & 11.36 & 40.91\\
 & W2V10 		& 12.50 & 35.23\\
 & OUR-APP 		& 0 & 54.55 \\ \hline
\multirow{3}{*}{200} 
 & RANDOM 		& 11.36 & 47.73\\
 & TF-IDF/PCA & 10.22 & 51.14\\
 & W2V10 		& 11.36 & 45.45\\
 & OUR-APP 		& 09.09 & 67.04 \\ \hline
\multirow{3}{*}{400} 
 & RANDOM 		& 07.95 & 62.50\\
 & TF-IDF/PCA & 06.81 & 64.77\\
 & W2V10 		& 09.09 & 60.23\\
 & \textbf{OUR-APP} & \textbf{05.68} & \textbf{78.40} \\ 
\hline
\end{tabular}
\caption{With neural attention}
\label{tab:lstm2}
\end{center}
}
\end{table}
\end{minipage} 
\end{tabular}

In Table ~\ref{tab:lstm2} the tests were run on the LSTM encoder-decoder architecture with attention mechanism. In next table we show examples of inputs and outputs generated by our model.

\begin{center}
{\scriptsize
\begin{tabular}{ |l|l| }
\hline
Input & Output (Without IRI Prefix to simplify) \\ \hline
\multirow{2}{*}{how many rivers are there in idaho ?}
&  SELECT (COUNT(?A) AS ?QTD) \{ ?idaho p:river ?A \\
& FILTER (regex(str(?idaho), "idaho", "i")) . \} \\\hline
\multirow{2}{*}{show major cities in colorado ?}
& SELECT ?A \{ ?colorado p:city ?A FILTER (regex(str(?colorado), "colorado", "i")) .\\ 
& ?A r:type p:Major . \} \\\hline
what are the cities of the state & SELECT ?B \{ ?A p:city ?B . \{ SELECT ?A \{ ?A r:type p:State . ?A p:highest\_point ?B .  \\
with the highest point ? & ?B p:height ?height . \}  ORDER BY DESC(?height) LIMIT 1 \} \} \\
\hline
\end{tabular}
}
\end{center}

Regarding the task of transforming the natural language into SPARQL, we compared our work with \cite{alagha2015using} and \cite{kaufmann2006querix}. In the first paper, the authors also use the Geo880 dataset and through linguistic analysis identify elements of natural language and generate RDF triples. In the second paper, the main strategy of the authors was to try to associate triples of natural language with RDF triples. As can be seen in the Table ~\ref{tabsparql2}, we obtained better results in the tests with dataset Geo880.
\vspace{-3mm}
\begin{table}[H]
\begin{center}
\begin{scriptsize}
\begin{tabular}{c|r|r|}
\cline{2-2}
& Accuracy  \\ \cline{1-2}
\multicolumn{1}{ |l| }{ \cite{alagha2015using}} & 58.61\\ \cline{1-2}
\multicolumn{1}{ |l| }{ Querix \cite{kaufmann2006querix} } & 77.67 \\ \cline{1-2}
\multicolumn{1}{ |l| }{Our method} & 78.40  \\ \cline{1-2}
\end{tabular}
\caption{Natural Language to SPARQL comparisons}
\label{tabsparql2}
\end{scriptsize}
\end{center}
\end{table}
\vspace{-8mm}
Although we mention the work \cite{wang2007panto}, that also makes use of the Geo880 dataset, we do not make the comparison with it because it does not use the original set of Geo880 queries in their tests. For the same reason, we also disregard one of the results of Querix \cite{kaufmann2006querix}. With respect to the \textbf{syntactical errors}, we can see in our tests that the better the model in general, the lower the error rate of syntax. We also propose as future work to develop a model that the generated sentence has syntactic guarantee.
\vspace{-4mm}
\section{Related Work}
\label{sec:relatedwork}

In this section we discuss three related works that are related to different aspects of our model. The first, \cite{dong2016language}, is related to the task of mapping natural language sentences to the logical form. The second \cite{alagha2015using} and third \cite{kaufmann2006querix}, are related with the task of translating natural language for SPARQL.

\cite{dong2016language} present a general method based on an attention-enhanced sequence- to-sequence model. They encode input sentences into vector representations using recurrent neural networks, and generate their logical forms by conditioning the output on the encoding vectors. The model is trained in an end-to-end fashion to maximize the likelihood of target logical forms given the natural language inputs. Although they do not deal with SPARQL, their approach uses a neural network attention-based structure similar to ours.

The work in  \cite{alagha2015using} associates phrases in natural language with RDF triples, as in our approach. Through a linguistic analysis, their model extracts relations and associates them to triples. They also generate SPARQL scripts, however using first an intermediate format. Then, with the help of an ontology, a SPARQL query is generated after identifying the targets and modifiers of the query. That is developed using Arabic as the source language, however a comparison with ours can be made as both works also use information extraction from a syntactic tree.

The Querix \cite{kaufmann2006querix} employs a statistical approach. Given a query, the system consists of parsing, removing important elements and then looking for triples that are related to the elements of the query. Querix works as a component and can be adapted in any application. It is based on clarification of dialogues, so when there is ambiguity the system asks the user to decide. All these works were evaluated using the GEO 880 dataset.

\section{Discussion and Future Work}

The purpose of this work was to explore artificial neural network resources in the development of a model that may be able to translate from natural language to SPARQL. The choice of the OWL Ontology and the SPARQL language as the target language is due to the fact that we are interested in practical applications. Among the advantages of using artificial neural networks, we can highlight the fact that we do not need linguistic knowledge nor do we depend on the development of complex grammars.

In addition to dealing with SPARQL, we propose in this work a representation of the target language lexicon that according to the results was a good approach. This representation is only possible because we use the concept of Matching of terms oriented by the alignment table that is generated by the mechanism of neural attention. Moreover, we can highlight as main future works: To perform tests with different datasets and to guarantee of correct syntax in query generation.

\section*{Acknowledgment}

This work was developed with the support of the National Council for Scientific and Technological Development (CNPq). Marcelo Finger was partly supported by Fapesp projects 2015/21880-4 and 2014/12236-1 and CNPq grant PQ 306582/2014-7.

\bibliographystyle{sbc}
\bibliography{paper-2017}

\end{document}